\documentclass[10pt, a4paper]{article}
\usepackage{lrec2022} 
\usepackage{multibib}
\newcites{languageresource}{Language Resources}
\usepackage{graphicx}
\usepackage{tabularx}
\usepackage{soul}
\usepackage{amsmath}
\usepackage{booktabs}
\usepackage{svg}
\usepackage{float}
\usepackage{adjustbox}
\usepackage[hyphens]{url}
\usepackage{makecell}

\usepackage{titlesec}
\titleformat{\section}{\normalfont\large\bfseries\center}{\thesection.}{1em}{}
\titleformat{\subsection}{\normalfont\SmallTitleFont\bfseries\raggedright}{\thesubsection.}{1em}{}
\titleformat{\subsubsection}{\normalfont\normalsize\bfseries\raggedright}{\thesubsubsection.}{1em}{}
\renewcommand\thesection{\arabic{section}}
\renewcommand\thesubsection{\thesection.\arabic{subsection}}
\renewcommand\thesubsubsection{\thesubsection.\arabic{subsubsection}}

\usepackage{epstopdf}
\usepackage[utf8]{inputenc}

\usepackage{hyperref}
\usepackage{xstring}

\usepackage{color}

\usepackage{csquotes}

\title{Multitask Learning for Grapheme-to-Phoneme Conversion of Anglicisms in German Speech Recognition}

\name{Julia Pritzen$^{1,2}$, Michael Gref$^{~1}$, Dietlind Zühlke$^2$, Christoph Schmidt$^1$}

\address{$^1$Fraunhofer Institute for Intelligent Analysis and Information Systems (IAIS), Germany\\
$^2$TH Köln - University of Applied Sciences, Cologne, Germany\\
\{julia.pritzen, michael.gref, christoph.andreas.schmidt\}@iais.fraunhofer.de\\ dietlind.zuehlke@th-koeln.de}

\abstract{
Anglicisms are a challenge in German speech recognition. Due to their irregular pronunciation compared to native German words, automatically generated pronunciation dictionaries often include faulty phoneme sequences for Anglicisms. In this work, we propose a multitask sequence-to-sequence approach for grapheme-to-phoneme conversion to improve the phonetization of Anglicisms. We extended a grapheme-to-phoneme model with a classifier to distinguish Anglicisms from native German words. With this approach, the model learns to generate pronunciations differently depending on the classification result. We used our model to create supplementary Anglicism pronunciation dictionaries that are added to an existing German speech recognition model. Tested on a dedicated Anglicism evaluation set, we improved the recognition of Anglicisms compared to a baseline model, reducing the word error rate by 1~\% and the Anglicism error rate by 3~\%. We show that multitask learning can help solving the challenge of Anglicisms in German speech recognition.\\
\newline
}

\begin{document}

\maketitleabstract

\section{Introduction}
\label{sec:intro}
Pronunciation dictionaries of hybrid automatic speech recognition (ASR) and Text-to-Speech (TTS) systems, which contain all word forms, are usually generated automatically using a grapheme-to-phoneme (G2P) model trained on a baseline dictionary. Being trained on one language only, monolingual G2P models often struggle with loanwords of foreign heritage since their pronunciation is derived from their source language. In the German language, the use of English loanwords (Anglicisms) has been steadily increasing \cite{burmasova_2010_anglicisms}, resulting in a ratio of 4.5~\% Anglicisms in spontaneous speech \cite{hunt_anglicisms}. German G2P models show higher error rates for Anglicisms due to their irregular pronunciation compared to native German words \cite{milde_2017_multitask}.

Currently, a lot of ASR research has moved away from hybrid ASR systems, moving more towards end-to-end modeling. However, for real-world applications in non-English languages, we believe there is still a need for improved phonetic pronunciation. In end-to-end approaches, adaptation to the non-canonical pronunciation of words, such as Anglicisms, would mainly be achievable via fine-tuning. This requires sufficient amounts of transcribed speech with Anglicisms, which, in general, is not always available. The pronunciation lexicons for training G2P systems are usually more common and easier to obtain.

Generating phoneme sequences with a G2P model trained on German data for Anglicisms can lead to wrong pronunciations. Looking at the Anglicism \enquote{Whistleblower}, a conventional Sequitur G2P system trained on PHONOLEX Core \citelanguageresource{phonolex} generates the pronunciation /v I s t l e: p l o 6/ by applying the pronunciation rules it learned from the German training data, but the correct pronunciation according to Duden is /v I s l b l O U6/\footnote{\url{https://www.duden.de/rechtschreibung/Whistleblower\#aussprache}}. 

We propose a multitask learning (MTL) approach to solve this problem, detecting Anglicisms in a second classification task parallel to G2P conversion. In MTL, the human concept of inductive transfer is applied to a machine learning model \cite[p.19]{caruana_1997_multitask}. By detecting Anglicisms based on the input sequence, the model generates phoneme sequences differently depending on the Anglicism classification result. To inspect this approach, we implemented a sequence-to-sequence G2P model with an additional classification task for Anglicism detection to create pronunciations for a list of automatically crawled Anglicisms. We evaluated the influence of the dictionary on the automatic speech recognition performance using a dedicated Anglicism test set.

\section{Related Work}
\label{sec:related}
Sequence-to-sequence (Seq2Seq) models are a deep neural network approach for handling sequences of unknown dimensions. They make use of LSTM cells \cite{LSTM_hochreiter}, a more complex kind of RNN cell. While a traditional RNN struggles with long-term dependencies, an LSTM model is able to handle information over long periods of time by controlling, forgetting, and passing information through the cell states. In the first LSTM implementation for automatic machine translation, \newcite{trans_seq2014} used an LSTM network as an encoder to obtain a fixed dimensional vector representation of an input sequence. As a decoder, they used an LSTM network conditioned on the input sequence to extract the output sequence from the vector. This way, a model is trained that maps source language input sequences to target language output sequences. While experimenting, they discovered that reversing the order of the input sequence elements positively influences the performance. If the LSTM reads the input sentence in reverse, many short-term dependencies in the data are introduced, which simplifies the optimization problem.

\begin{figure*}[htb]
  \centering
  \includesvg[width=0.9\textwidth]{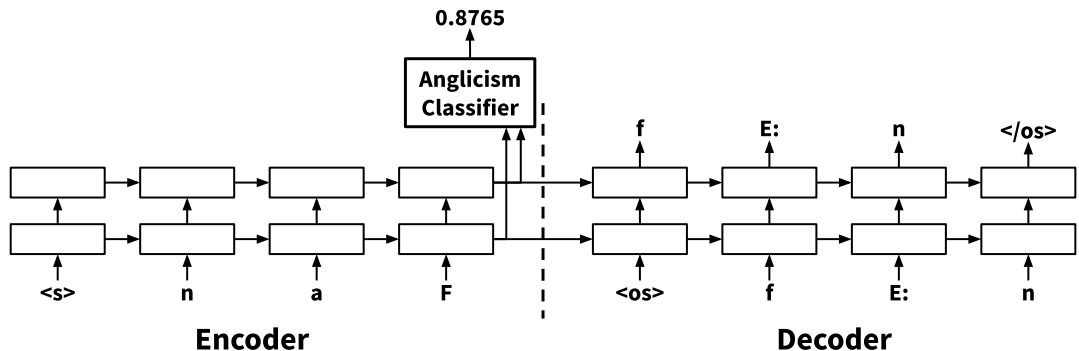}
  \caption{Seq2Seq G2P model with additional Anglicism classification task processing the input sequence \textlangle Fan\textrangle~with phoneme output /f E: n/ in BAS-SAMPA notation (adapted from \protect\newcite{yao_seq2seq}). The input sequence is read in reverse.}
  \label{fig:seq2seq_mtl_model}
\end{figure*}

\newcite{yao_seq2seq} applied the method of \newcite{trans_seq2014} to the G2P task, giving similar results as traditional joint-sequence models. Tested on the CMUdict dataset, their two-layered encoder-decoder LSTM model showed a phoneme error rate (PER) of 7.63~\% and a word error rate (WER) of 28.61~\%, performing slightly worse than the baseline Sequitur G2P \cite{Bisani_2008_g2p} (PER: 5.88~\%, WER: 24.53~\%). However, the Seq2Seq G2P approach enables novel applications in the field of deep learning and will likely benefit from future improvements in recurrent neural network architectures \cite{milde_2017_multitask}.

Seq2Seq models can be used to train a multilingual G2P model due to their ability of joint learning of alignments and grapheme-phoneme translations \cite{sokolov_2019_amazon}. \newcite{sokolov_2019_amazon} built a multilingual Seq2Seq G2P model utilizing transfer learning to improve the performance on foreign words from different languages. They determine a language ID value in advance based on the previously known distribution of the word from different given language models. This language ID vector is an additional input to the model, concatenated to the attention vector. They trained both a monolingual and a multilingual G2P model respectively for 18 languages. The approach improved prediction accuracy compared to monolingual models for low-resource languages. However, we assume it does not apply to the Anglicism problem tackled in our work. The English stem words in Anglicisms are often “Germanized” and usually do not occur with this spelling in any other language.

Another option that the neural network approach of Seq2Seq G2P models facilitates is MTL. First introduced by \newcite{caruana_1993}, MTL is an approach of modeling the human concept of inductive transfer to a machine learning model. When humans are confronted with a new problem, they use the skills and information they already learned for related problems in the past. As opposed to single-task learning, where every task is learned separately from each other, MTL allows learning multiple tasks in parallel. The tasks of an MTL model have to be related to each other. Related tasks provide an inductive bias, making the model learn more general representations. 

Hard parameter sharing is the most common approach for MTL. The input layers are shared between all tasks, while the output layers are kept task-specific. Hard parameter sharing reduces the risk of overfitting because the model is forced to find a more general representation that fits all tasks instead of only one. Having multiple tasks also helps to differentiate between relevant and irrelevant features since each task provides evidence for the feature's relevance. This shifts the models' focus towards those genuinely essential features.

MTL fits tasks in the field of natural language processing well since text contains various cues that can be helpful for multiple tasks simultaneously. \newcite{milde_2017_multitask} built three multilingual Seq2Seq G2P models utilizing MTL, training them simultaneously on a German and English G2P task using the PHONOLEX and CMUdict data sets. Their models used character embeddings as encoder inputs and phoneme embeddings as decoder inputs. They added a language marker at the start of each input sequence for classifying the source language. The encoder was built as a stacked bi-directional LSTM to represent past and future dependencies. Overall, the MTL models did not outperform the baseline Sequitur G2P model. The MTL models were also tested on specific word groups inside the German PHONOLEX test set, including a set of English loanwords. The Sequitur G2P model outperformed the MTL models within this word group, even though the MTL combinations additionally contained an English G2P task.

\section{Proposed Approach}
\label{sec:approach}

While the MTL approach by \newcite{milde_2017_multitask} did not show improvements for English loanwords, it inspired us to use MTL for solving the challenge of Anglicism pronunciations in the German language. Since Anglicisms are of English heritage, their different linguistic features (grapheme combinations) compared to native German words can be an indicator for detecting Anglicisms and hence applying different pronunciation rules. Classifying grapheme sequences as Anglicisms can help the model understand that Anglicisms are pronounced differently than native German words, resulting in different phoneme conversions.

Our proposed approach is shown in Figure \ref{fig:seq2seq_mtl_model}. As basis, we used the encoder-decoder-LSTM model by \newcite{yao_seq2seq} with two layers. The functioning of the model is illustrated in the figure with an example. The encoder LSTM reads the reversed input grapheme sequence \enquote{\texttt{\textlangle s\textrangle} n a F}, where \texttt{\textlangle s\textrangle} indicates the beginning of the sequence. After the last hidden layer activation, the decoder LSTM is initialized. It produces \enquote{\texttt{\textlangle os\textrangle} f E: n} as phoneme prediction of the input sequence and uses \enquote{f E: n \texttt{\textlangle /os\textrangle}} as the output sequence. \texttt{\textlangle os\textrangle} and \texttt{\textlangle /os\textrangle} indicate the start and end of the output sequence, respectively. The encoder LSTM represents the entire input sequence in the hidden layer activities which are used as initial activities of the decoder. Working as a language model, the decoder LSTM uses the past phoneme sequence to predict the next phoneme. It stops predicting after outputting \texttt{\textlangle /os\textrangle}. 

An Anglicism classifier is added as a second task in the Seq2Seq G2P model. Utilizing hard parameter sharing, the output vectors of the encoder are used as input for both the decoder and the binary Anglicism classification task. The model optimizes both tasks by combining their losses. Looking at Figure \ref{fig:seq2seq_mtl_model}, the grapheme sequence \textlangle Fan\textrangle~is processed by the encoder that passes the output to both the decoder and the Anglicism classification task. Based on the encoder output, the decoder generates the pronunciation while the classification task asserts the probability for the grapheme sequence being an Anglicism.

\section{Experimental Setup}
\label{sec:setup}
To test the viability of our proposed approach, we applied it to a list of Anglicisms to generate pronunciations that will be used as a supplementary pronunciation dictionary in a German hybrid ASR model. The model corresponds to the source model used by \newcite{gref_2019} with a slightly different language model. It was trained on the GER-TV1000h corpus \citelanguageresource{gertv1000h}. Depending on the model performance on a specific Anglicism test set, we can assess the potential for more extensive applications.

\begin{table*}[t]
	\centering
	\adjustbox{max width=\textwidth}{%
		\begin{tabular}{@{}llrrrrrrrr@{}}
			\toprule
			& & & & \multicolumn{2}{c}{\textbf{G2P Task}} & \multicolumn{4}{c}{\textbf{Anglicism Classification Task}} \\
			\cmidrule(l{1em}r{1em}){5-6} \cmidrule(l{1em}r{1em}){7-10}
			\thead{G2P\\Model} & \thead{Data Source \& Specifics}   & \thead{Epochs} & \thead{Iter. /\\Epoch} & \thead{PER}    & \thead{WER}  & \thead{Accu.} & \thead{Prec.}  & \thead{Recall}   & \thead{F1} \\
			\midrule
			$\text{MTL}_{\text{Base}}$ & PHONOLEX core & 7 & 2498 & \textit{5.68}   & \textit{24.43}  & \textit{98.03}  & \textit{0.00} & \textit{0.00}   & \textit{0.00} \\
			\midrule
			$\text{MTL}_{\text{Wiki}}$ & $\text{MTL}_{\text{Base}}$ + Wiktionary Anglicisms & 6 & 2845 & 8.63    & 30.89   & 91.24   & 80.69   & 54.26   & 64.89 \\
			$\text{MTL}_{\text{WL}}$ & $\text{MTL}_{\text{Wiki}}$ + weighed losses ($\alpha = 0.7$) & 7 & 2845 & \textbf{7.87}   & \textbf{28.03}  & \textbf{92.42}  & 86.71   & 58.14   & 69.61 \\
			$\text{MTL}_{\text{DS}}$ & $\text{MTL}_{\text{Wiki}}$ + downsampled data & 16 & 806 &  11.21   & 39.63   & 88.66  & \textbf{90.30}  & \textbf{86.63}   & \textbf{88.43} \\
			\bottomrule
		\end{tabular}
}
\caption{Selected MTL models with their training information as well as their G2P task and Anglicism classification task evaluation metrics. The PER and WER measures are based on a fixed percentage split of the training data. For $\text{MTL}_{\text{Base}}$, the precision, recall and F1 score values are 0.00~\% because they did not yield any positive classifications. All metrics and error measures are given in percent.}
\label{tab:mtl_model_metrics}
\end{table*}

\begin{table}[htb]
\centering
\begin{tabular}[t]{@{}lrrr@{}}
\toprule
\thead{G2P Model} & \thead{PHONOLEX\\Core\\PER (\%)} & \thead{Wiktionary\\Anglicisms\\PER (\%)} \\ \midrule
Sequitur    & \textbf{2.59} & 17.11 \\
Seq2Seq     & 5.13          & 19.80 \\
$\text{MTL}_{\text{Base}}$     & 5.68          & 25.72 & \\
$\text{MTL}_{\text{Wiki}}$     & 7.41          & 16.92 & \\
$\text{MTL}_{\text{WL}}$     & 6.63          & 15.98 & \\
$\text{MTL}_{\text{DS}}$     & 12.69         & \textbf{11.57} \\
 \bottomrule
\end{tabular}
\caption{PER values of the PHONOLEX Core and Wiktionary Anglicism validation data for the baseline and MTL G2P models. The Seq2Seq model corresponds to the basis of all MTL models and is trained with the same data as $\text{MTL}_{\text{Base}}$. The values for PHONOLEX Core show the general performance on German data, while the results for the Wiktionary Anglicism data show the specific performance for Anglicisms.}
\label{tab:per_mtl}
\end{table}

\subsection{Datasets}
\label{ssec:datasets}
To create an Anglicism word list, we derived 11,839 Anglicisms from Wiktionary's list of German Anglicisms \citelanguageresource{wiki_anglicisms} and Pseudo-Anglicisms \citelanguageresource{wiki_pseudoanglicisms} as well as the VDS Anglizismenindex \citelanguageresource{vds_2020}. Additionally, inflections of the contained words were crawled from the Wiktionary website, expanding the word list to 18,967 entries. We used PHONOLEX core as training data for the G2P model, using 62,427 entries as train set and 3,000 entries as the validation set. We classified the lexicon entries based on the Anglicism word list. Since PHONOLEX core only contained 2.22~\% words classified as Anglicism in the train set, we derived 9,802 additional Anglicism pronunciations from Wiktionary and added them to the training data. With this data added to PHONOLEX core, the train set contained 71,102 entries, including 10,063 Anglicisms (16.11~\%). The validation set contained 3,457 entries, including 516 Anglicisms (17.20~\%). Based on this data, we created an additional downsampled data set that offers a 50/50 class balance between Anglicisms and non-Anglicisms, resulting in 20,126 entries in the train set and 1,032 entries in the validation set.

For the evaluation of the resulting ASR models, we created a test set (\enquote{Anglicisms 2020}) including segments with Anglicism usage. The data was derived from newscasts, business \& technical talks, and videos containing colloquial speech, resulting in 1.3~h of audio data. We annotated the audio data using ELAN \cite{elan_2020}. For evaluating the specific performance of Anglicism recognitions, we flagged every Anglicism in the annotations.
Of 14,028 total words, 1,362 were marked as Anglicisms (9.71~\%). To ensure that our approach would not negatively influence the general performance of native German words, we also used two in-house test sets representing typical German broadcast use cases as control groups. The audio data for those test sets were derived from 0.94~h of television segments (\enquote{German Broadcast 2020}) and 0.99~h of radio interviews containing spontaneous speech (\enquote{Challenging Broadcast 2018}).

\subsection{Implementation}
\label{ssec:implementation}
For our Seq2Seq G2P model, we rebuilt the encoder-decoder LSTM from \newcite{yao_seq2seq}. Like in \cite{yao_seq2seq}, the model training was set up using 500-dimensional projection and hidden layers and applying back-propagation through time. We used beam search to generate the phoneme sequence during decoding, selecting the hypothesis sequence with the highest posterior probability as the decoding result. We used a batch size of 25 for our data set as it performed best on the validation data among various configurations. The order of the training sequences was randomly permuted in each epoch. We used an adaptive learning rate of $0.007$ that was halved throughout training when no improvements in the validation loss were observed within the last five checks. An early stopping mechanism set in when the learning rate dropped below $0.00001$.

\begin{table*}[t]
\centering
\begin{tabular}[t]{@{}lrrrrr@{}}
\toprule
& \multicolumn{3}{c}{\textbf{Anglicisms 2020}} & & \\
\cmidrule(l{1em}r{1em}){2-4}
\thead{ASR Model} & \thead{WER (\%)} & \thead{AER (\%)} & \thead{Recognized\\Anglicisms} & \thead{German\\BC 2020\\WER (\%)} & \thead{Challenging\\BC 2018\\WER (\%)} \\ \midrule
Baseline \cite{gref_2019}    & 15.80             & 39.50     & 824   & \textbf{6.56}     & 10.84 \\
Sequitur    & 15.76             & 39.35     & 826   & \textbf{6.56}     & \textbf{10.82} \\
Seq2Seq     & 15.75             & 39.28     & 827   & \textbf{6.56}     & 10.91 \\
$\text{MTL}_{\text{WL}}$      & \textbf{15.65}    & \textbf{38.33}    & \textbf{840}  & 6.57  & 10.86 \\
$\text{MTL}_{\text{DS}}$      & 15.67             & 38.40     & 839   & 6.60              & 10.90 \\
\midrule
Wav2Vec2                     & 15.69 & 42.07 & 789   & 9.34 & \textbf{9.48} \\
\bottomrule
\end{tabular}
\caption{Evaluation results for the baseline and MTL models. As baseline, we used a German ASR model based on the source model in \protect\newcite{gref_2019}. All other models extend the baseline with an additional Anglicism pronunciation dictionary based on the respective G2P approach. For the Anglicism 2020 test set, an additional Anglicism error rate (AER) is reported, indicating the percentage of correctly recognized Anglicisms. The German broadcast test sets served as control groups. The last row additionally shows the results of a Wav2Vec2 model. Since our Wav2Vec2 model could not handle hyphens, all hyphens in the reference transcripts were mapped to whitespaces to simulate a fair comparison to the other models.}
\label{tab:wer_asr_long}
\end{table*}

\begin{table*}[htb]
\centering
\begin{tabular}{@{}lllll@{}}
\toprule
          & \textbf{Sequitur} & \textbf{Seq2Seq} & \textbf{$\text{MTL}_{\text{WL}}$} & \textbf{$\text{MTL}_{\text{DS}}$} \\ \midrule
Boomers   & b u: m 6 s        & b u: m 6 s       & b u: m 6 s      & b u: m 6 s      \\
Brownie   & b r aU n i:       & b r o v n i:     & b r aU n j @    & b r o v i:      \\
Cosplay   & k O s p l e:      & k O s p l e:     & k O s p l E I   & k O s p l e:    \\
spreadet  & s p r E tS E t    & S p r i: d @ t   & S p r i: d @ t  & S p r i: d @ t  \\
used      & j u: s t          & z e: t           & Q u: z @ t      & Q aU s d        \\
virgin    & v I6 g I n        & v I6 g I n       & f I6 g I n      & v I6 g I n      \\ \bottomrule
\end{tabular}
\caption{Example entries from the Anglicism pronunciation dictionaries of the compared ASR models in BAS-SAMPA notation. While some pronunciations are similar (e.g.~\enquote{Boomers}) others show strong differences (e.g.~\enquote{used}). For some words, none of the G2P models was able to produce a suitable pronunciation. For example, the pronunciation for  \enquote{virgin} would be /v~I6~dZ~I~n/ where the grapheme \textlangle g\textrangle~is pronounced as a /dZ/, but all models chose the phoneme /g/ for their result.}
\label{tab:anglicism_dict}
\end{table*}

We added a binary classification task as an additional task after the encoder step, transforming the single task encoder-decoder LSTM model into an MTL model. The classifier consists of two hidden layers and an output layer. We combined the 500-dimensional cell state and cell output resulting from the encoder and used them as input for the classification task. The first hidden layer was a 1,000-dimensional linear layer with a 100-dimensional output. ReLU was used as the activation function. We applied a dropout of 0.2 to prevent overfitting. The second hidden layer was a 100-dimensional linear layer with equal output using PReLU with a constant $\alpha = 1$ as the activation function. The output layer was a 100-dimensional linear layer with one output neuron. We used the sigmoid function to get an output value between 0 and 1. The closer the output value is to 1, the more likely the word is an Anglicism.

For the G2P decoder, we used LogSoftmax as the output activation function in the output layer. The loss was calculated with the negative log-likelihood since it usually goes in combination with Softmax. We calculated the classifier loss with the binary cross-entropy as this fits best with a binary classifier with an output value between 0 and 1. To optimize on both tasks, we combined both losses to one total loss value in the training and validation phase:
\begin{equation}
    \text{Total Loss} = \text{Decoder Loss} + \text{Classifier Loss}
    \label{eq:sum_loss}
\end{equation}
Based on an input grapheme sequence, the resulting MTL Seq2Seq G2P model was able to both generate a corresponding phoneme sequence as well as classify whether the input sequence is considered an Anglicism.

\section{Evaluation and Results}
\label{sec:evaluation}
Table \ref{tab:mtl_model_metrics} shows the metrics of our MTL G2P models. Model $\text{MTL}_{\text{Base}}$ being trained on PHONOLEX core shows how the class imbalance in the training data made the model only choose negative classifications. Adding the Wiktionary Anglicism pronunciations to the train data in model $\text{MTL}_{\text{Wiki}}$ helped getting viable classification results. For model $\text{MTL}_{\text{WL}}$, we decided to alter the loss summation by including an additional $\alpha$-parameter in the loss summation to weight the tasks accordingly:
\begin{equation}
    \text{Total Loss} = \alpha \cdot \text{Decoder Loss} + (1 - \alpha) \cdot \text{Classifier Loss}
    \label{eq:weighed_loss}
\end{equation}
We chose $\alpha = 0.7$ to put more influence on the decoder task, which led to both improved PER and WER as well as improved classification metrics based on the validation set. We trained model $\text{MTL}_{\text{DS}}$ on the downsampled data set with equal loss summation (see Equation \ref{eq:sum_loss}). This setup shows the highest PER and WER of all models but also the best precision and recall values. While this was most likely caused by the higher number of entries with positive Anglicism classifications in the training data, we were interested in how the performance of this differently trained classifier would affect the ASR results.

To get a more universal assessment of the MTL G2P models performance, we compared them to two traditional monolingual German G2P models as the baseline. We used the Seq2Seq G2P model that our MTL models are based on, but without classification task, and the German Sequitur G2P model currently used at Fraunhofer IAIS. We used the same PHONOLEX Core training and validation data as for the MTL models but without added Anglicism pronunciations from Wiktionary. Table \ref{tab:per_mtl} shows the PER results of the baseline and MTL G2P models. They were calculated on the PHONOLEX Core validation set (3.000 entries) and the Wiktionary Anglicism pronunciations from the MTL model's validation sets (516 entries) that were not included in either G2P models' training data. Similar to the results of \newcite{milde_2017_multitask}, all Seq2Seq G2P models showed increased PER values for the PHONOLEX Core validation data compared to Sequitur G2P. Looking at the MTL models, we observed a decreasing Anglicism PER with an increasing Anglicism ratio in the training data. Overall, the additional classification task seemed to worsen the performance on native German words while it helped to generate more accurate Anglicism pronunciations.

For evaluating the actual performance in an ASR setup, we chose models $\text{MTL}_{\text{WL}}$ and $\text{MTL}_{\text{DS}}$ to create a supplementary Anglicism pronunciation dictionary used in an ASR model. Based on an existing ASR model, we added the resulting Anglicism pronunciations to the pronunciation dictionary. We created two dedicated ASR models for the two MTL G2P models with this method.
To compare the performance of the generated Anglicism pronunciations with those of traditional G2P models, we created two more ASR models by generating Anglicism pronunciations with a Sequitur and a Seq2Seq G2P model. We created the Anglicism pronunciation dictionaries based on a list of 18,967 Anglicisms that was derived from Wiktionary and VDS Anglizismenindex (see Section \ref{ssec:datasets}).

Along with the baseline ASR model that did not include an additional Anglicism dictionary, we tested these models on the Anglicism ASR test set \enquote{Anglicisms 2020}. To make sure the results for native German words are not affected by the additional Anglicism pronunciations, we also tested the models on two typical German broadcast (BC) test sets \enquote{German BC 2020}, and \enquote{Challenging BC 2018}. We measured the WER to determine the overall performance of the added pronunciations. To specifically evaluate the performance of Anglicisms, we measured an Anglicism error rate (AER) by flagging every Anglicism in the test set \enquote{Anglicisms 2020}. Based on the number of all Anglicisms in the test set, the AER represents the ratio of wrongly recognized Anglicisms.

Table \ref{tab:wer_asr_long} shows the evaluation results of the Anglicism test set. While the WER and AER of all models showed improved results compared to the baseline model due to the additional Anglicism pronunciations in the dictionary, both MTL models outperformed the non-MTL models. $\text{MTL}_{\text{WL}}$ showed the best WER, decreasing the WER of the baseline model by relatively 1~\% and the AER by 3~\%, with 16 more Anglicisms being recognized. The results for test sets \enquote{German BC 2020} and \enquote{Challenging BC 2018} show that the WERs of $\text{MTL}_{\text{WL}}$ only increased by an absolute 0.01~\% and 0.02~\%, respectively, which shows that the additional Anglicism pronunciations did not significantly impact the performance of typical German applications.
Given that Anglicisms only account for a small fraction of German spoken language \cite{hunt_anglicisms}, our approach successfully improved the performance of Anglicisms in German ASR without negatively impacting the performance of typical German applications.

Considering the recent rise of end-to-end models, we additionally did a benchmark on a Wav2Vec2 model \cite{baevski2020wav2vec}. The model was implemented using the Hugging Face Transformers library \cite{huggingface_transformers}. We used Facebook's XLSR-Wav2Vec2 \cite{facebook_wav2vec} fine-tuned on the German Common Voice dataset \citelanguageresource{commonvoice}, provided by \newcite{jonatas} as \enquote{Wav2Vec2-Large-XLSR-53-German}, as our base model and further fine-tuned it on the GER-TV1000h corpus \citelanguageresource{gertv1000h}. To make the results comparable, we also applied the same language model that was used in our other models.

During our tests, we noticed that the Wav2Vec2 model was not able to output hyphens. Since it is relevant to our use cases, we usually consider both casing and hyphen differences when calculating the WER. As the results of the other models did contain hyphens, we tried to resolve this disadvantage of Wav2Vec2 by mapping all hyphens in the reference transcripts to whitespaces.

With this mapping applied, Table \ref{tab:wer_asr_long} shows the evaluation results of the Anglicism test set as well as the two control sets. For test set \enquote{Anglicisms 2020}, the Wav2Vec2 model showed a WER of 15.69~\%, which at first sight compares reasonably well to the MTL model results. However, taking a closer look at the recognized Anglicisms, we calculated an AER of 42.07~\%, exceeding the baseline model by an absolute 2.57~\%, and hence showed the highest AER among all tested models. Looking at the results for the two control groups, we assume that the Wav2Vec2 model performs better for German in general but has more problems with Anglicisms, which is reflected in the increased AER. We expected a better performance since the XLSR-Wav2Vec2 was (i.a.) trained on 557~h of English audio, which we thought might positively influence the recognition of Anglicisms. We need to continue experimenting with end-to-end ASR models to get a more reasonable comparison and improve our results with further fine-tuning.

For a more qualitative evaluation, we looked at the entries of the resulting supplementary Anglicism pronunciation dictionaries. Table \ref{tab:anglicism_dict} shows eight example Anglicism entries with their generated pronunciations from the respective models. While we observed cases where the generated pronunciations were similar, for example, in \enquote{Boomers} and \enquote{Cosplay}, there were also cases where the phoneme sequences noticeably differed from each other as in \enquote{Brownie} and \enquote{used}.
For some pronunciations, we observed that no model was able to generate a proper pronunciation, e.g.~\enquote{spreadet} and \enquote{virgin}. Since this concerned all models, it might be caused by insufficient training data that is missing or under-representing certain grapheme-phoneme-combinations that the models struggle to learn.

Looking more into the training data, we found that some Anglicisms were potentially misclassified. As the Anglicism classification of the 65,427 entries in the PHONOLEX core data was done automatically based on an Anglicism list, all words that were not included have not been declared as an Anglicism. We must further evaluate and modify the training data to improve the Seq2Seq model's learning process.

We plan to extend the Anglicism list by using a similar approach to \cite{anglicism_corpus} to detect more Anglicisms in the training data. Here, Coats created an Anglicism corpus based on social media data by applying linguistic rules. After generating potential Anglicisms with a rule-based approach, he cross-checked them against German Twitter data to determine which Anglicisms exist and are, in fact, used in every day (written) language. With this method, we can extend our Anglicism list and potentially gather more existing pronunciations from sources like Wiktionary to further improve our training data.

\section{Conclusions}
\label{sec:conclusions}
In this work, we propose a multitask sequence-to-sequence training to enhance the generation of Anglicism pronunciations by a German G2P model. With our approach, we improved the Anglicism recognition results by generating and adding Anglicism pronunciations to the ASR model's pronunciation dictionary. While positively influencing the Anglicism recognition results for our dedicated Anglicism test set, our approach did not noticeably disturb the performance of other test sets representing typical use cases in the broadcast domain. By only modifying the pronunciation dictionary of an existing ASR model, the improvements on the Anglicism test set (WER $-1 \%$ relative and AER $-3 \%$ relative) show that our approach has the potential to tackle the challenge of Anglicisms in German ASR.

Since our approach uses only phonemes of the German phoneme set, the resulting pronunciations can be added to a pronunciation lexicon of an existing ASR system without adapting the acoustic model. Another advantage of using only the German phoneme set is that the resulting phoneme sequences refer to the German pronunciation of Anglicisms, rather than the reuse of foreign phonemes that a German speaker may not be able to pronounce depending on their language skills. These "Germanized" pronunciations are more realistic with respect to real-world applications. In addition to ASR systems, which we have focused on in this work, we assume that our approach can also help improve the pronunciation of loan words in German TTS systems.

The limited Anglicism data was an issue for creating the Seq2Seq training data. With more Anglicism pronunciations, the classification and decoding tasks might improve further due to more training material. Also, possible misclassifications in the training data could have negatively impacted the learning phase of the classification task. By extending the list of Anglicisms for automatically classifying the training data and additional manual checks, the classification results and the generation of Anglicism pronunciations could be further improved.

While we used the same parameter configurations for all Seq2Seq G2P models to better compare the results in this publication, we plan on optimizing individual setups for the different models in the future. We also want to experiment with the tuning criteria, e.g.~focusing more on the classification results to better deal with the class imbalance. After further optimizing our MTL models, we plan on using the approach in a bigger scope by generating an entire pronunciation dictionary instead of only adding supplementary Anglicism entries.

We also plan on looking more into end-to-end models. The reported WER for the Wav2Vec2 model only resulted from our first experiments with this technology and had the additional restriction of a language model. We are looking forward to improving the results and finding out if end-to-end models could also be a possible solution for the problem of Anglicisms and other loanwords in the German language.

\section{Acknowledgments}
\label{sec:acknowledgments}
We thank Tilo Himmelsbach from Fraunhofer IAIS for his work on the Wav2Vec2 model.

\section{Bibliographical References}\label{reference}

\bibliographystyle{lrec2022-bib}
\bibliography{lrec2022_mtl_anglicisms}

\section{Language Resource References}
\label{lr:ref}
\bibliographystylelanguageresource{lrec2022-bib}
\bibliographylanguageresource{languageresource}

\end{document}